\documentclass[fleqn,10pt]{wlscirep}
\usepackage{float} 
\usepackage{tabularx}
\usepackage{multirow} 
\usepackage[utf8]{inputenc}
\usepackage[T1]{fontenc}
\usepackage{graphicx}
\usepackage{subcaption}
\usepackage{tabularx} 
\title{A Comparative Performance Analysis of Classification and Segmentation Models on Bangladeshi Pothole Dataset}

\author[1,*]{Antara Firoz Parsa}
\author[1]{S. M. Abdullah}
\author[1]{Anika Hasan Talukder}
\author[1]{Md. Asif Shahidullah Kabbya}
\author[1]{Shakib Al Hasan}
\author[2]{Md. Farhadul Islam}
\author[2]{Jannatun Noor}

\affil[1]{Department of Computer Science and Engineering, School of Data and Sciences, BRAC University, Dhaka, Bangladesh}
\affil[2]{Computing for Sustainability and Social Good (C2SG) Research Group, Department of Computer Science and Engineering, United International University, Dhaka, Bangladesh}

\affil[1]{antara.firuz.parsa@g.bracu.ac.bd}
\affil[1]{s.m.abdullah@g.bracu.ac.bd}
\affil[1]{anika.hasan.talukder@g.bracu.ac.bd}
\affil[1]{asif.shahidullah.kabbya@g.bracu.ac.bd}
\affil[1]{shakib.al.hasan@g.bracu.ac.bd}
\affil[2]{farhadulfuad324@gmail.com}
\affil[2]{jannatun@cse.uiu.ac.bd}

\affil[*]{Corresponding Author}

\begin{abstract}

The study involves a comprehensive performance analysis of popular classification and segmentation models, applied over a Bangladeshi pothole dataset, being developed by the authors of this research. This custom dataset of 824 samples, collected from the streets of Dhaka and Bogura performs competitively against the existing industrial and custom datasets utilized in the present literature. The dataset was further augmented four-fold for segmentation and ten-fold for classification evaluation. We tested nine classification models (CCT, CNN, INN, Swin Transformer, ConvMixer, VGG16, ResNet50, DenseNet201, and Xception) and four segmentation models (U-Net, ResU-Net, U-Net++, and Attention-Unet) over both the datasets. Among the classification models, lightweight models namely CCT, CNN, INN, Swin Transformer, and ConvMixer were emphasized due to their low computational requirements and faster prediction times. The lightweight models performed respectfully, oftentimes equating to the performance of heavyweight models. In addition, augmentation was found to enhance the performance of all the tested models. The experimental results exhibit that, our dataset performs on par or outperforms the similar classification models utilized in the existing literature, reaching accuracy and f1-scores over 99\%. The dataset also performed on par with the existing datasets for segmentation, achieving model Dice Similarity Coefficient up to 67.54\% and IoU scores up to 59.39\%. 

\end{abstract}

\begin{document}

\flushbottom
\maketitle
%
%
\thispagestyle{empty}

\vspace{2mm}

\section*{Introduction}

Potholes are defective holes or cavities on road surfaces. Potholes are a major factor behind road accidents in Bangladesh\cite{Islam2021}. These holes pose great danger for pedestrians and motorcyclists alongside contributing to other traffic related accidents\cite{Hoque2011}. Potholes are also responsible for great economic loss that occurred due to vehicular damage\cite{Adid2024}. As per a survey conducted between November 2019 and March 2020, the overall district roads condition in Bangladesh were observed to be deteriorating. According to another study, 23\% roads of Rajshahi City were found in failure condition which is counterintuitively considered among the good examples of non-capital district roads in Bangladesh\cite{Hasan2020}.
 \vspace{1mm}

Accurate detection of potholes can play a crucial role in the road safety scenario of Bangladesh, potentially preventing major accidents. The use of deep learning model architectures can facilitate the reliable detection of potholes and cracks on roads. Machine learning models require robust datasets for training and validation. There is a substantial gap of the Bangladeshi pothole dataset in the existing literature. Moreover, there was absence of any public pothole datasets in the context of Bangladesh as per our findings. 
\vspace{1mm}

To address this gap, the authors of the research have proposed and developed a dataset in the Bangladeshi context. The custom dataset contains 824 images captured from the streets of Dhaka and Bogura. Various pre-processing methods have been applied to create the dataset. It was further augmented to enhance data diversity. Selected machine learning models were then applied on both datasets. To illustrate performance, we conducted both classification and segmentation which are the two fundamental methods of evaluating performance in computer vision. We tested nine classification models (CCT, Custom CNN, Custom INN, Swin Transformer, ConvMixer, VGG16, ResNet50, DenseNet201, and Xception) and four segmentation models (U-Net, ResU-Net, U-Net++, and Attention U-Net) over both the datasets. We have additionally addressed another literature gap by putting emphasis on the performance of lightweight models (CCT, CNN, INN, Swin Transformer, ConvMixer), alongside the traditional pre-trained heavyweight models for classification. Lightweight models have low computational requirements and faster prediction times, making them suitable for real-time detection tasks like pothole detection. This research provides a detailed perspective of performance comparison of lightweight models against the popular traditional heavyweight models, applied over our dataset in order to mitigate the literature gap. Additionally, the existing literature also lacked a comprehensive evaluation of the effects of augmentation in classification and segmentation tasks. We have tested all 13 models on the augmented dataset for both classification and segmentation and have provided a more comprehensive, accurate, and mature conclusion regarding the effects of dataset augmentation on machine learning models. Last but not least, the research compares performance of our dataset against existing literatures in order to evaluate effectiveness of the proposed dataset.
\vspace{1mm}

Overall we can summarize the paper’s key contributions as of following-
\begin{itemize}
    \item We proposed and created a pothole dataset in context of Bangladesh containing images from the streets of Dhaka and Bogura.
\end{itemize}
\begin{itemize}
    \item We augmented the dataset and compressively compared its performance with the raw dataset across 13 machine learning models (including results of both classification and segmentation) in order to evaluate the influence of augmentation on deep learning models.
\end{itemize}
\begin{itemize}
    \item Alongside, we separately analyzed the performance of lightweight models in contrast to traditional pre-trained heavyweight models among our tested architectures, with the aim to present a new outlook and consideration for the lightweight models due to their better appropriation for real-time detection tasks like pothole detection.
\end{itemize}
The research paper is further divided into the following segments: The ‘Related Works’ segment involves summary of the existing research in this field. The ‘Methods’ segment describes the chronological research methodology including Dataset Collection and Processing, Data Augmentation, and description of the Applied Models. In the ‘Results’ segment, we outline the comprehensive performance analysis of classification and segmentation across both datasets. Furthermore, the ‘Discussion’ segment highlights and provides deeper insights of the major research findings. Lastly, the ‘Conclusion’ segment concludes the study.
\vspace{2mm}

\section*{Related Works}

Bhutad et al., 2022\cite{Bhutad2022} discuss about different dataset collection strategies that were maintained to improve the quality of their pothole dataset. The paper tells that any different weather conditions that are not present in the dataset may lead to inaccurate detection of that weather condition. Their dataset aimed to provide accurate results for the summer and rainy seasons. Additionally, the pothole images were captured from both top and side views, improving the diversity and characteristics of the data. The capturing device wasthe  Samsung Galaxy A22. The final dataset included 10 videos and 8484 images. Whang et al., 2023\cite{Whang2023} provide a comprehensive analysis of data collection methods emphasizing on measures on data cleaning and data sanitization etc.
\vspace{1mm}

Pramanik et al., 2021\cite{Pramanik2021} tested VGG16 and ResNet50 over a dataset of 1490 images collected by the authors. 80\% data were used for training and 20\% for testing. The applied image size for VGG16 was 224x224 and for ResNet50, 256x256. After data collection, the dataset was augmented which enhanced picture samples and solved the problem of overfitting. The accuracies obtained from VGG-16 and ResNet50 were 96.31\% and 98.66\% respectively. Arjapure et al. 2020\cite{Arjapure2020} tested the performance of CNN, DenseNet201, ResNet152, ResNet50, ResNet50V2, ResNet152V2, InceptionV3, and InceptionResNetV2 over a dataset of 838 images utilizing 86\% training and 14\% testing data. DenseNet201 and InceptionResNetV2 performed better than other models with an accuracy of 89.66\%. Veturi et al. 2023\cite{Veturi2023} tested the classification performance of various deep-learning models where ResNet50 had the highest accuracy of 95\%.
\vspace{1mm}

Shijie et al., 2017\cite{Shijie2017} emphasized on both supervised and unsupervised approaches of data augmentation. Unsupervised approaches involve methods like flipping, cropping, rotation, color jittering, shifting, and noise addition, etc. Supervised approaches involve methods like Generative Adversarial Networks (GANs) and their variations. Such methods generate completely new images based on corresponding images from the raw dataset, but require very high computational powers. The influence of data augmentation was evaluated on classification tasks on a proposed architecture based on CNN, namely AlexNet. The utilized datasets were- CIFAR10 and a part of ImageNet. Augmentation was found to improve the performance compared to the non-augmented dataset. The performance enhanced more using larger augmentation sample sizes. Additionally, the paper shows that combining several augmentation methods (e.g. flipping + rotating, flipping + WGAN) can perform marginally better than individual techniques. The paper also claims that augmentation work more effectively on smaller training sets. In addition, Ucar et al., 2022\cite{Ucar2021} utilize shifting, rotating, and flipping as the unsupervised augmentation methods for their study. The study reveals that augmentation may oftentimes generate faulty images, however, it guaranties improvement of prediction accuracy at all cases.  Furthermore, Shorten et al., 2019\cite{Shorten2019} experimented with ten augmentation methods and concluded that augmentation showcases small datasets as large datasets and helps avoid overfitting.

\section*{Methods}
\vspace{1mm}

\subsection*{Dataset Collection and Processing}
\vspace{1mm}

\begin{figure}[h!]
    \centering
    \begin{subfigure}[b]{0.3\textwidth}
        \centering
        \includegraphics[width=\textwidth]{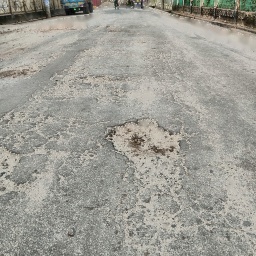}
        \caption{Sample 1 (Pothole)}
        \label{fig:image1}
    \end{subfigure}
    \hspace{3mm}
    \begin{subfigure}[b]{0.3\textwidth}
        \centering
        \includegraphics[width=\textwidth]{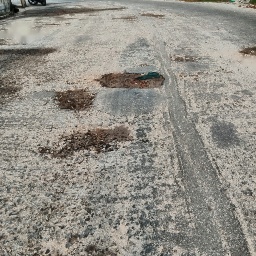}
        \caption{Sample 2 (Pothole)}
        \label{fig:image2}
    \end{subfigure}
    \hspace{3mm}
    \begin{subfigure}[b]{0.3\textwidth}
        \centering
        \includegraphics[width=\textwidth]{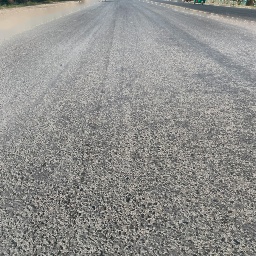}
        \caption{Sample 3 (Non-Pothole)}
        \label{fig:image3}
    \end{subfigure}
    \caption{Classification Dataset Samples.}
    \label{fig:sidebyside1}
\end{figure}

\begin{figure}[h!]
    \centering
    \begin{subfigure}[b]{0.25\textwidth} 
        \centering
        \includegraphics[width=\textwidth]{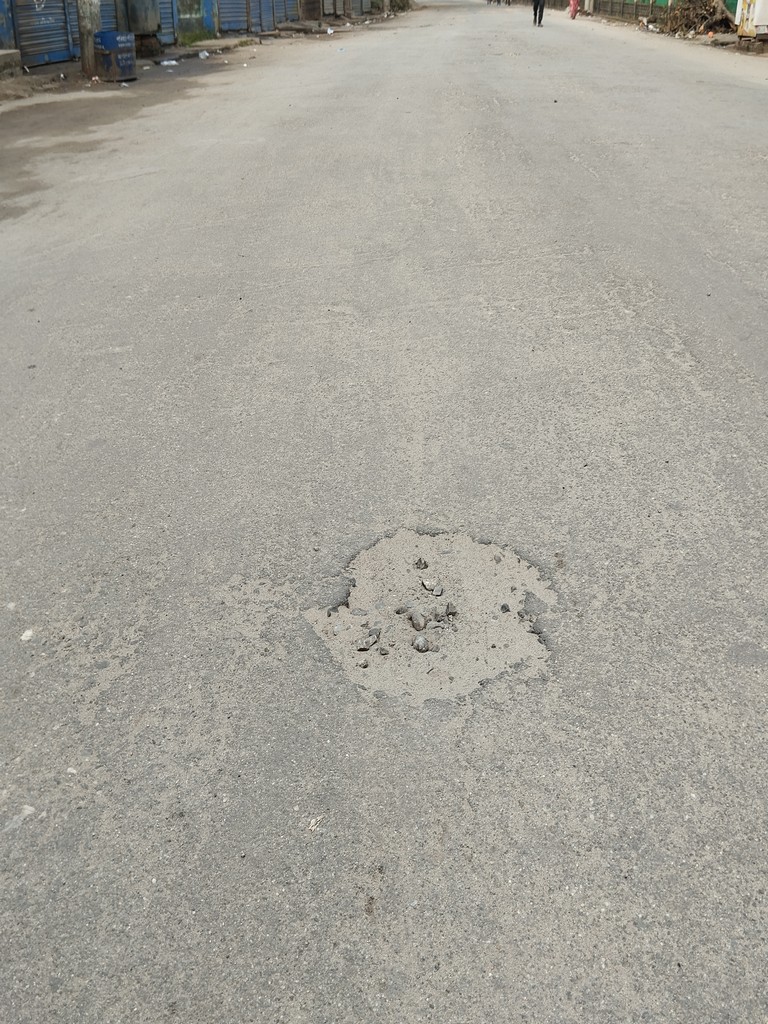}
        \caption{Sample 1 (Raw)}
        \label{fig:image1}
    \end{subfigure}
    \hspace{5mm} 
    \begin{subfigure}[b]{0.25\textwidth}
        \centering
        \includegraphics[width=\textwidth]{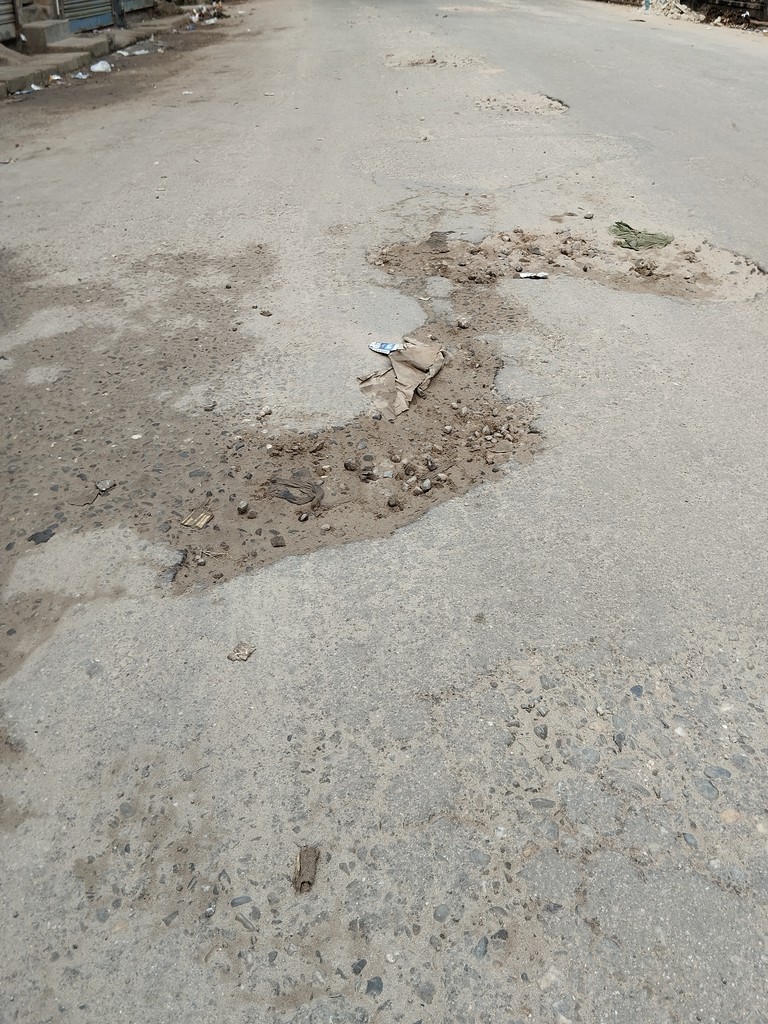}
        \caption{Sample 2 (Raw)}
        \label{fig:image2}
    \end{subfigure}
    \hspace{5mm}
    \begin{subfigure}[b]{0.25\textwidth}
        \centering
        \includegraphics[width=\textwidth]{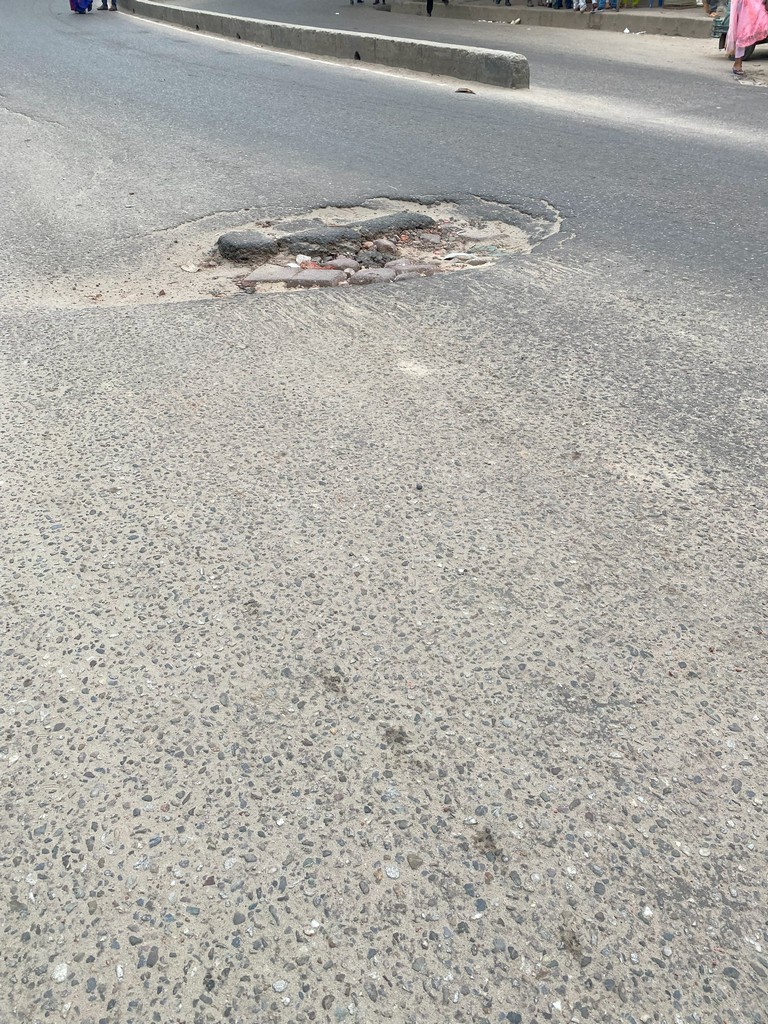}
        \caption{Sample 3 (Raw)}
        \label{fig:image3}
    \end{subfigure}

    \vspace{5mm} 

    \begin{subfigure}[b]{0.25\textwidth}
        \centering
        \includegraphics[width=\textwidth]{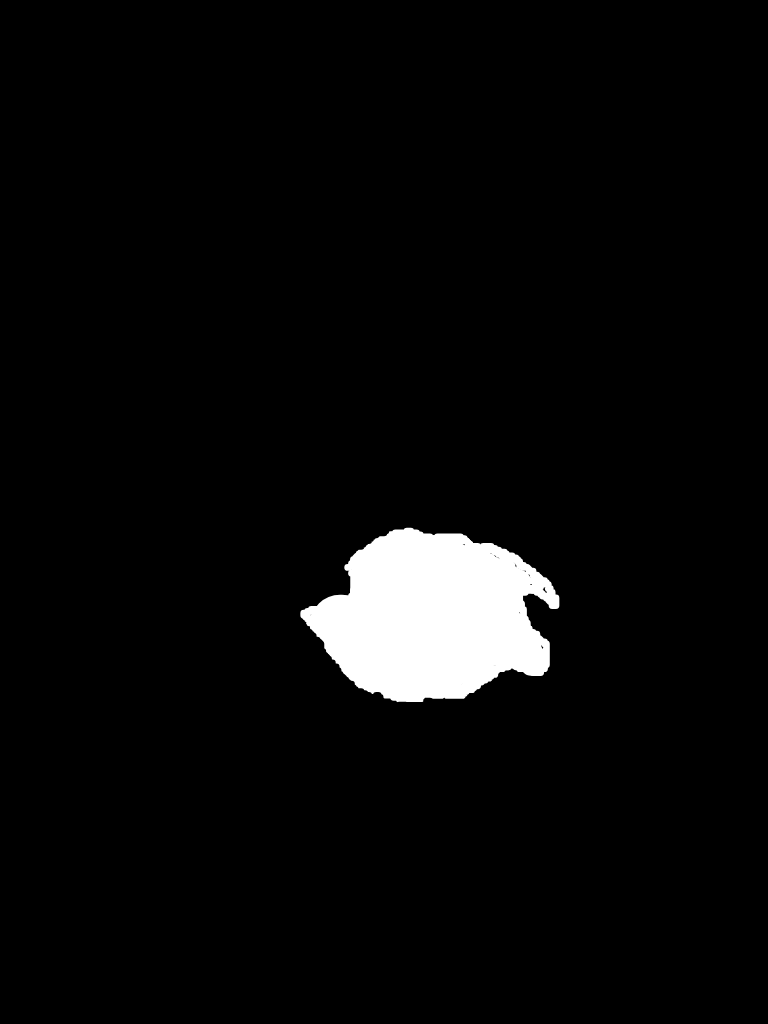}
        \caption{Sample 1 (Mask)}
        \label{fig:image4}
    \end{subfigure}
    \hspace{5mm}
    \begin{subfigure}[b]{0.25\textwidth}
        \centering
        \includegraphics[width=\textwidth]{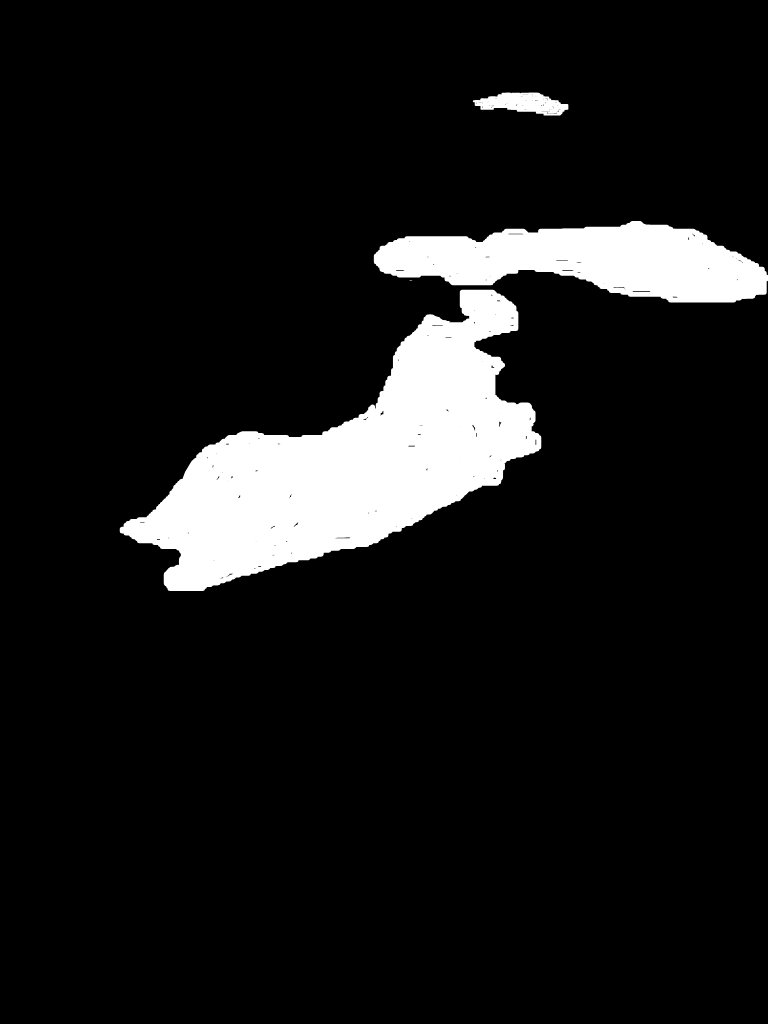}
        \caption{Sample 2 (Mask)}
        \label{fig:image5}
    \end{subfigure}
    \hspace{5mm}
    \begin{subfigure}[b]{0.25\textwidth}
        \centering
        \includegraphics[width=\textwidth]{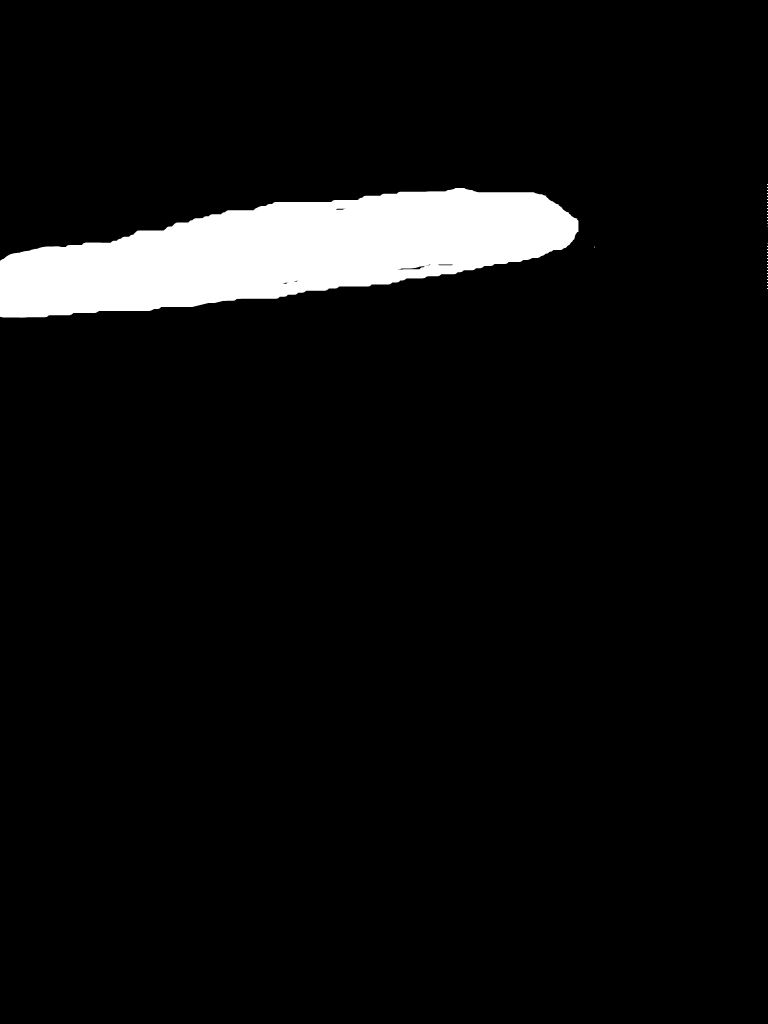}
        \caption{Sample 3 (Mask)}
        \label{fig:image6}
    \end{subfigure}

    \caption{Segmentation Dataset Samples.}
    \label{fig:sidebyside2}
\end{figure}

The authors accumulated road samples of plain roads and potholes from the streets of Dhaka and Bogura. The dataset was collected and processed complying with dataset collection methodology enshrined in Bhutad et al., 2022\cite{Bhutad2022}. Among the collected images, a final refined dataset of 824 images was developed by only keeping the complying images. Image samples were captured on iPhone 11, iPhone XR, iPhone 6s Plus, and Xiaomi Redmi Note 12, ensuring acceptable image clarity. Unclear and duplicate images were excluded. Alongside, data cleaning was performed by excluding or blurring images that contained human faces or human-sensitive information, e.g. the car number plates, address plates, etc. Moreover, the samples were taken keeping an acceptable depth of view of the surrounding environment in order to better replicate the Bangladeshi street context. In addition, the Contrast Limited Adaptive Histogram Equalization (CLAHE) algorithm was applied to improve local contrast, making important features more visible, especially in under or over-exposed regions. Subsequently, the dataset samples were divided into two classes, Pothole and Non-Pothole. The two classes were divided into an almost similar ratio (437:387) in order to lessen class imbalance. The finalized dataset samples were then reformatted to ensure consistency in the dataset. The dataset image resolution was reduced to  $128\times128$ for classification, and $256\times256$ for segmentation, and samples were uniformly converted into JPEG format. The finalized raw dataset was then generated by allocating relevant images into the Pothole and Non-Pothole folders. A separate annotated dataset was fabricated for annotation. Figure \ref{fig:sidebyside1} and Figure \ref{fig:sidebyside2} showcase both the classification and segmentation samples from the proposed dataset.
\vspace{1mm}

\begin{figure}[ht]
\centering
\includegraphics[scale=0.6]{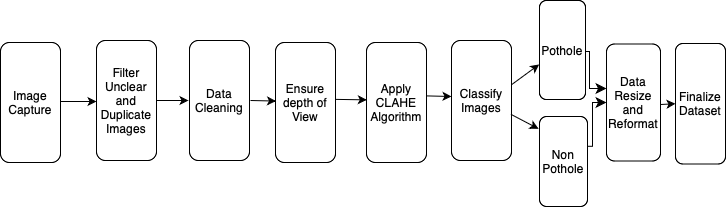}
\caption{Data Collection Methodology}
\label{fig:stream...}
\end{figure}
\vspace{1mm}

\vspace{1mm}

\subsection*{Dataset Augmentation}

Augmentation was performed over the raw dataset by utilizing methods including- left-right and top-down flipping; random brightness, random contrast, and random rotation; and zoom, skew tilt, and shear. The zoom was limited to 5\% and the rotation was limited to 22 degrees in order to prevent any pothole cut from the picture samples. Variations of brightness and contrast were set between 0.7 and 1.2 in order to prevent excessive brightness or contrast. For classification, the images were augmented ten-fold into 8240 $128\times128$ images, and for segmentation, images were augmented four-fold into 3296 $256\times256$ images. These augmentations introduced varied orientations and perspectives, enhancing the dataset's diversity and generalization compatibility. The ‘Results’ section of the paper indicates that augmentation improved the performance metric of nearly every model tested in this research. 
\vspace{1mm}

\subsection*{Applied Models}

Both the raw and the augmented dataset were tested on popular classification and segmentation models. 
\vspace{1mm}

\subsubsection*{Classification Models}
 
The tested classification models include five lightweight and four pre-trained traditional heavyweight models. Lightweight models include Custom CNN, Custom Involutional Neural Network (INN), Compact Convolutional Transformers (CCT), Convmixer and Swin Transformer. All lightweight models apply convolutional layers for spatial feature extraction of textures, edges, and complex shapes. Compared to fully connected layers, convolutional layers reduce the total number of parameters, propagating faster prediction times. Moreover, features like positional embeddings or patch embeddings in CCT and Swin Transformer encodes the spatial characteristics, enhancing structured and ordered feature recognition\cite{Jiang2022}. Along with that, all lightweight models also incorporate normalization layers like LayerNormalization or BatchNormalization which contribute to a smooth learning process by keeping activation in a controlled range. In addition, the heavyweight models include ResNet50, DenseNet201, VGG16, and Xception. All of them are pre-trained on the ImageNet dataset, contributing to reduced training times and enhanced ability to recognize new data\cite{Kornblith2018}. The training times of the heavyweight models were on par with the lightweight models, even surpassing models like Convmixer. The heavyweight models rely upon the transfer learning principle. The models keep the pre-trained weights unchanged by freezing the initial base network layers for which only new custom layers are trained. This way, model efficiency is enhanced, and computational requirements are reduced. It also minimizes over-fitting. This practice is beneficial when working with limited data, relevant to our raw dataset of 824 images. As a result, the heavyweight models perform comparatively better for the raw dataset compared to lightweight models, as illustrated in Table \ref{tab:example4}. Last but not least, 80\% of the data were utilized for training and 20\% for testing in classification.  
\begin{figure}[ht]
\centering
\includegraphics[scale=0.46]{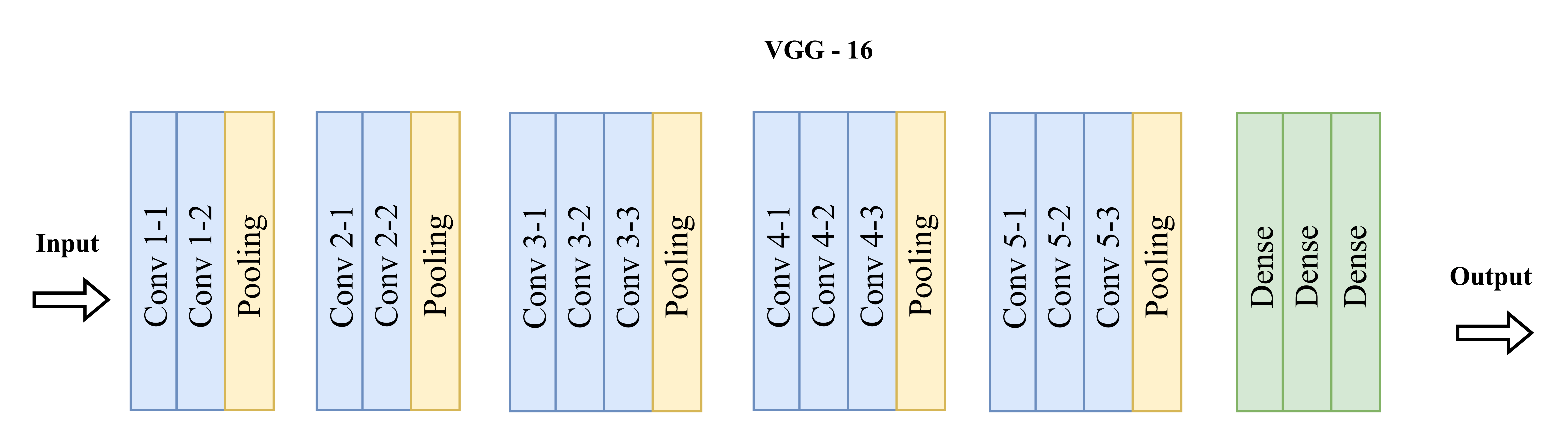}
\caption{VGG16 Architecture}
\label{fig:stream3}
\end{figure}
\vspace{1mm}

\subsubsection*{Segmentation Models}
 
The tested segmentation models include U-Net, U-Net++, Attention U-Net, and ResU-Net. All segmentation models include an encoder or contracting path and a decoder or expansive path\cite{Bousias2020}. Encoder downsamples an input and reduce spatial dimensions to compress hierarchical features for segmentation. Moreover, dropout layers are present in the encoder to prevent overfitting and enhance regularization and generalization capability. Decoder path on other hand performs the opposite operation by upsampling feature maps to recover the original dimensions. Additionally, Conv2DTranspose layers or UpSampling2D layers are utilized in the model architecture for proper reconstruction of spatial details for upsampling. Along with that skip connections is another characteristic of the utilized models which bridges encoder and decoder by merging downsampled encoder feature maps with upsampled decoder feature maps. For segmentation, 70\% of the data was utilized for training and 30\% for testing.

\begin{figure}[ht]
\centering
\includegraphics[scale=0.08]{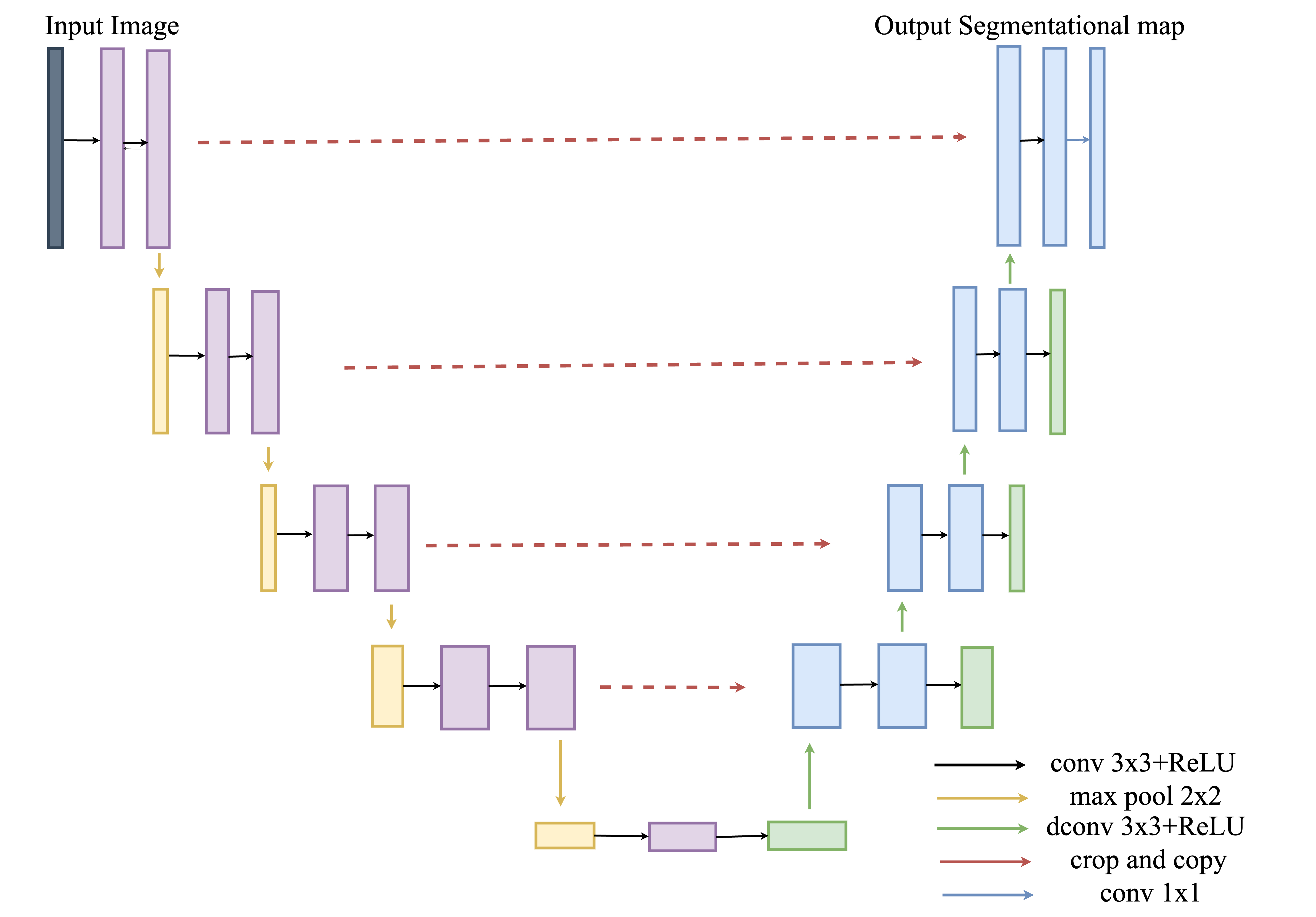}
\caption{U-Net Architecture}
\label{fig:stream3}
\end{figure}

\vspace{2mm}
\subsection*{Model Optimization} 
\vspace{1mm}

\subsubsection*{Optimization for Classification}
 
The models are formed for binary image classification to differentiate road types: ‘Non-Pothole’ and ‘Pothole’. Images are resized to 128x128 pixels. For consistency, pixel values are divided by 255 and the dataset is shuffled to ensure a balanced distribution of training and validation sets. During training, class weights are applied and calculated for addressing class imbalance. As a result, it mitigates the biasness for the more common class which ensures a balanced learning process. Therefore, balanced learning enhances classification accuracy. In addition, callback functions including EarlyStopping, ReduceLROnPlateau, and ModelCheckpoint have been integrated in the code to optimize model performance. To balance the speed with accuracy, $1e-4$ learning rate is used with the Adam optimizer. Batch size and image size of all the segmentation models are 16 and 128 respectively.

\begin{table}[ht]
\centering
\begin{tabular}{|l|l|l|l|l|l|}
\hline
Model & Epochs(Original Dataset) & Epochs(Augmented Dataset) & Learning Rate \\
\hline

Densenet201 & 75 &100 &  0.0001 \\
\hline
Xception & 81 &100 &  0.0001 \\
\hline
VGG16 & 41 & 59 &  0.0001 \\
\hline
ResNet50 & 100 & 100 & 0.0001 \\
\hline
INN & 100 &59 &  0.0001 \\
\hline
CNN & 100 &100 &  0.0001 \\
\hline
ConvMixer & 11 &100 &  0.0001 \\
\hline
CCT & 79 &100 &  0.0001 \\
\hline
Swin Transformer & 100 &100 &  0.0001 \\
\hline
\end{tabular}
\caption{\label{tab:example2}Model Code Optimization (Classification Models)}
\end{table}
\vspace{1mm}

\subsubsection*{Optimization for Segmentation}
 
A combined loss function is used on all four segmentation models merging the Binary Cross Entropy alongside Dice loss. Binary Crossentropy enhances pixel-wise classification for segmentation tasks, and Dice Loss works good for imbalanced data. For accurate performance evaluation, the Dice Coefficient and the Intersection over Union (IoU) metrics are added in the models. Dice Coefficient interprets similarity of reference and predication, and IoU reveals the accuracy of image segmentation\cite{Tang2024}. They also help monitor how effectively the actual mask matches with the predicted mask. All four segmentation models utilize the Adam optimizer  which should assist to adapt to gradient changes in complex segmentation tasks. Moreover, EarlyStopping and ReduceLROnPlateau callbacks are used on all models for improving coverage. EarlyStopping prevents over-fitting by stopping model-training. ReduceLROnPlateau optimizes the training progress by adjusting the learning rate based on validation performance which ultimately improves accuracy. Batch size and image size of all the segmentation models are 8 and 256 respectively.

\begin{table}[ht]
\centering
\begin{tabular}{|l|l|l|l|l|l|}
\hline
Model & Epochs(Original Dataset) & Epochs(Augmented Dataset) &  Learning Rate \\
\hline

U-Net & 150 &150 &  0.0001 \\
\hline
ResU-Net & 47 &27 &  0.0001 \\
\hline
U-Net++ & 50 &66 &  0.0001 \\
\hline
Attention U-Net & 82 &75 &  0.00001 \\
\hline
\end{tabular}
\caption{\label{tab:example2}Model Code Optimization (Segmentation Models)}
\end{table}

\vspace{2mm}

\section*{Results}
\vspace{1mm}

The following performance metrics are utilized in our research for classification- Model Accuracy, Precision, Recall, and F-Score in order to evaluate model prediction results over our dataset.

\[Test\;Accuracy = \frac{True\;Positives + True\;Negatives}{Total}\]

\[Precision = \frac{True\;Positives}{True\;Positives + False\;Positives}\]

\[Recall = \frac{True\;Positives}{True\;Positives + False\;Negatives}\]

\[F1-Score = \frac{2 \times(Precision \times Recall)}{Precision + Recall}\]

And, the following have been utilized for evaluating segmentation- Model Accuracy, Dice Coefficient, and Intersection over Union (IoU).

\[Test\;Accuracy = \frac{True\;Positives + True\;Negatives}{Total}\]

\[Dice\;Coefficient = \frac{2 \times |X \cap Y|}{|X|+|Y|} \]

\[IoU = \frac{|X \cap Y|}{|X \cup Y|} \]
\vspace{1mm}

Here, X and Y are two sets and the vertical bars denote the absolute value of the corresponding contents. Model accuracy refers to the proportion of correct predictions in comparison to total predictions. It is the simplest and the go-to performance metric in computer vision. Precision and recall evaluate the proportion of positive predictions in comparison to false positives and false negatives subsequently. F1-score on the other hand presents a balanced value of the precision and recall results. The dice coefficient and IoU measure the overlap between the predicted and ground truth segmentation. Dice is commonly used where pixel-level accuracy is crucial and IoU is widely used to evaluate object detection and instance segmentation performance. 
We evaluated the performance of our proposed dataset by comparing the performance of other custom datasets tested across similar models, utilized throughout existing literature, as presented in Table \ref{tab:related_work}. To illustrate, we tested 13 models and we searched for performance figures of the same models across the existing literature. The literature utilized different custom and industrial datasets. As the models were the same, the major differentiating performance factor here becomes the dataset as per the methodology. Thus, we were able to contrast the performance of particularly the datasets. Our dataset overall performed competently in comparison to existing industrial and custom datasets tested in the existing literature.

\vspace{1.5mm}

\begin{figure}[h!]
    \centering
    \begin{subfigure}[b]{0.29\textwidth}
        \centering
        \includegraphics[width=\textwidth]{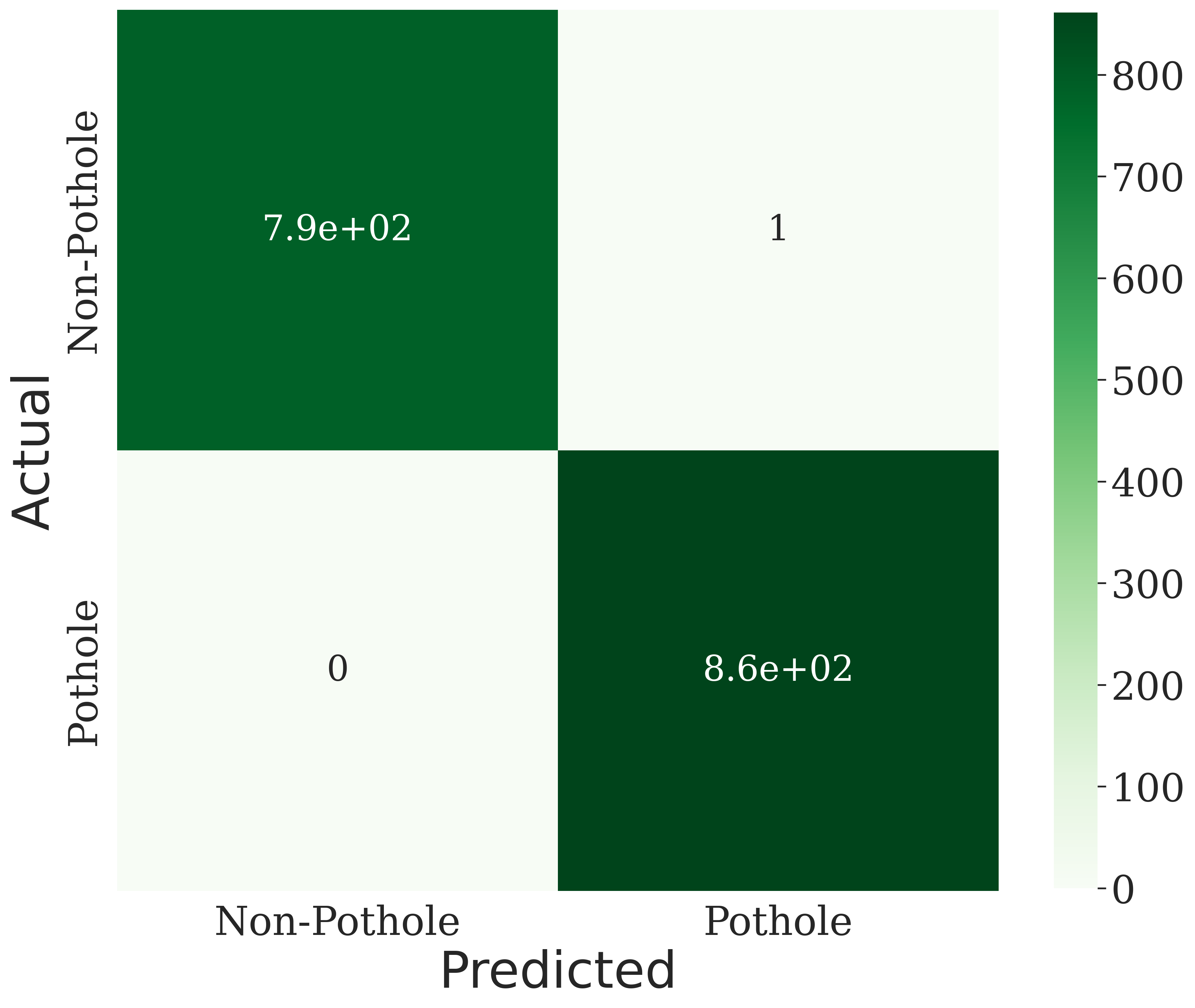}
        \caption{VGG16 (Augmented Dataset)}
        \label{fig:image1}
    \end{subfigure}
    \hspace{9mm} 
    \begin{subfigure}[b]{0.29\textwidth}
        \centering
        \includegraphics[width=\textwidth]{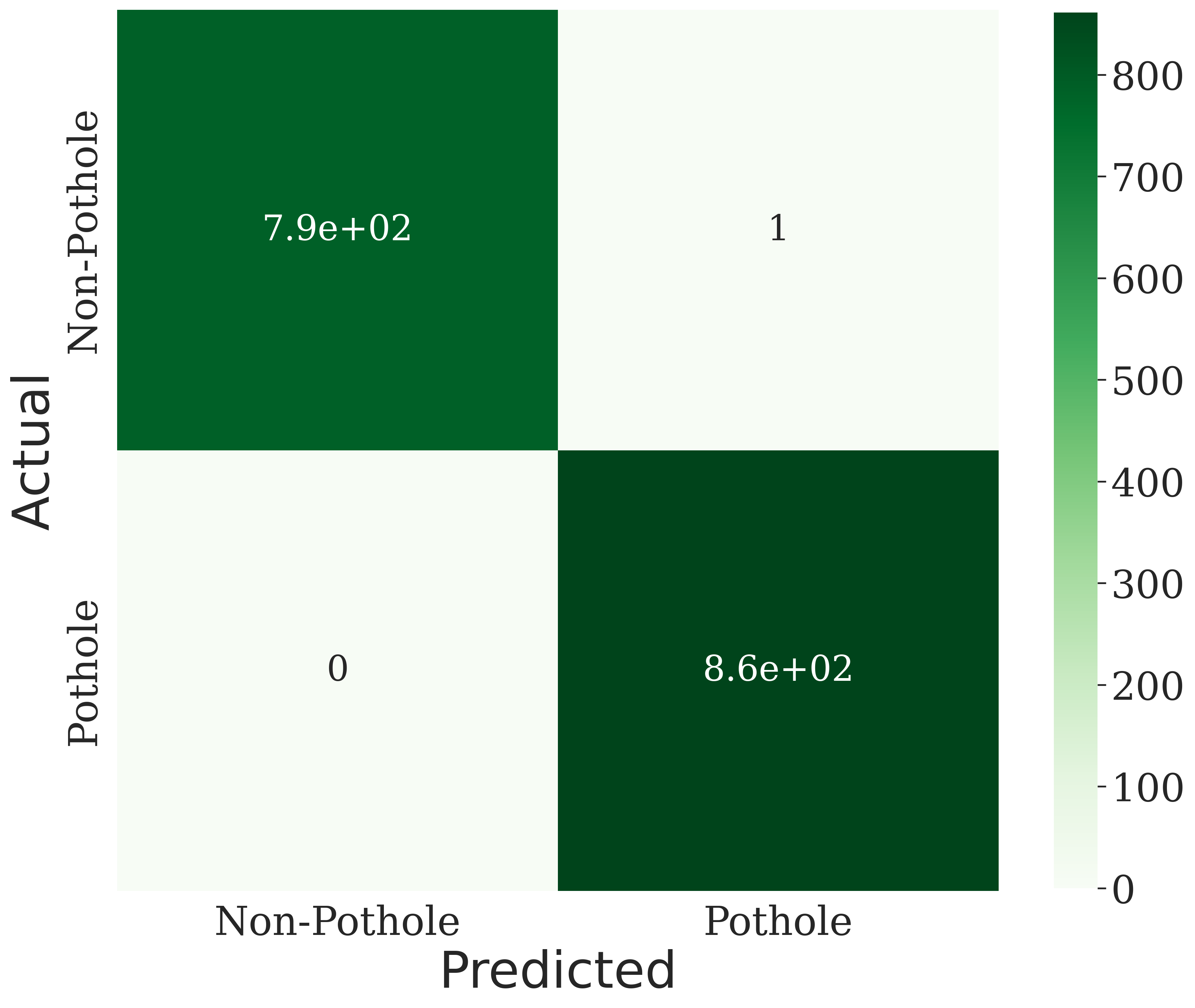}
        \caption{DenseNet201 (Augmented Dataset)}
        \label{fig:image2}
    \end{subfigure}
    \hspace{9mm} 
    \begin{subfigure}[b]{0.288\textwidth}
        \centering
        \includegraphics[width=\textwidth]{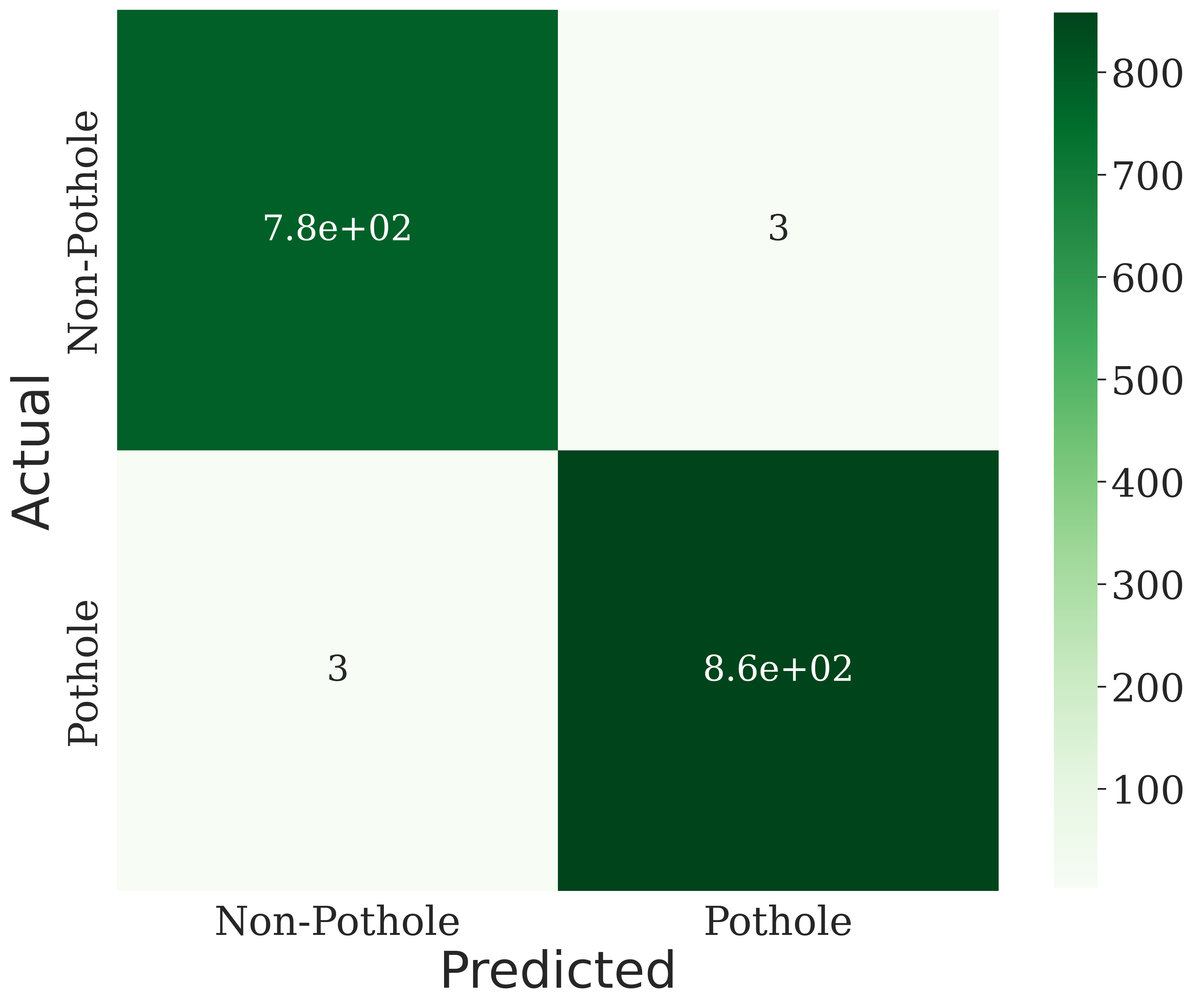}
        \caption{Xception (Augmented Dataset)}
        \label{fig:image2}
    \end{subfigure}
    \vspace{5mm}
    \begin{subfigure}[b]{0.28\textwidth}
        \centering
        \includegraphics[width=\textwidth]{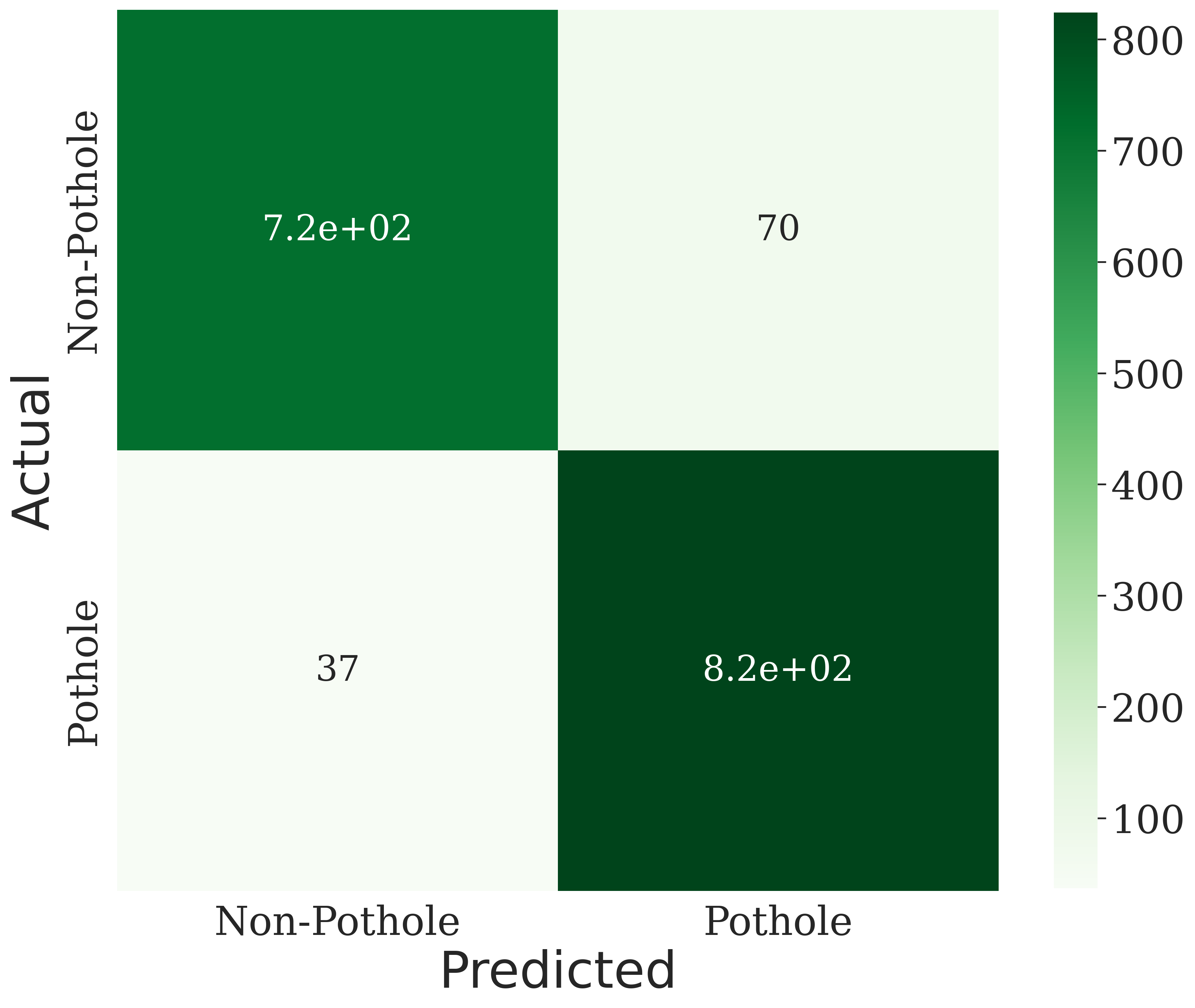}
        \caption{ReseNet50 (Augmented Dataset)}
        \label{fig:image2}
    \end{subfigure}
    \hspace{9mm} 
    \begin{subfigure}[b]{0.278\textwidth}
        \centering
        \includegraphics[width=\textwidth]{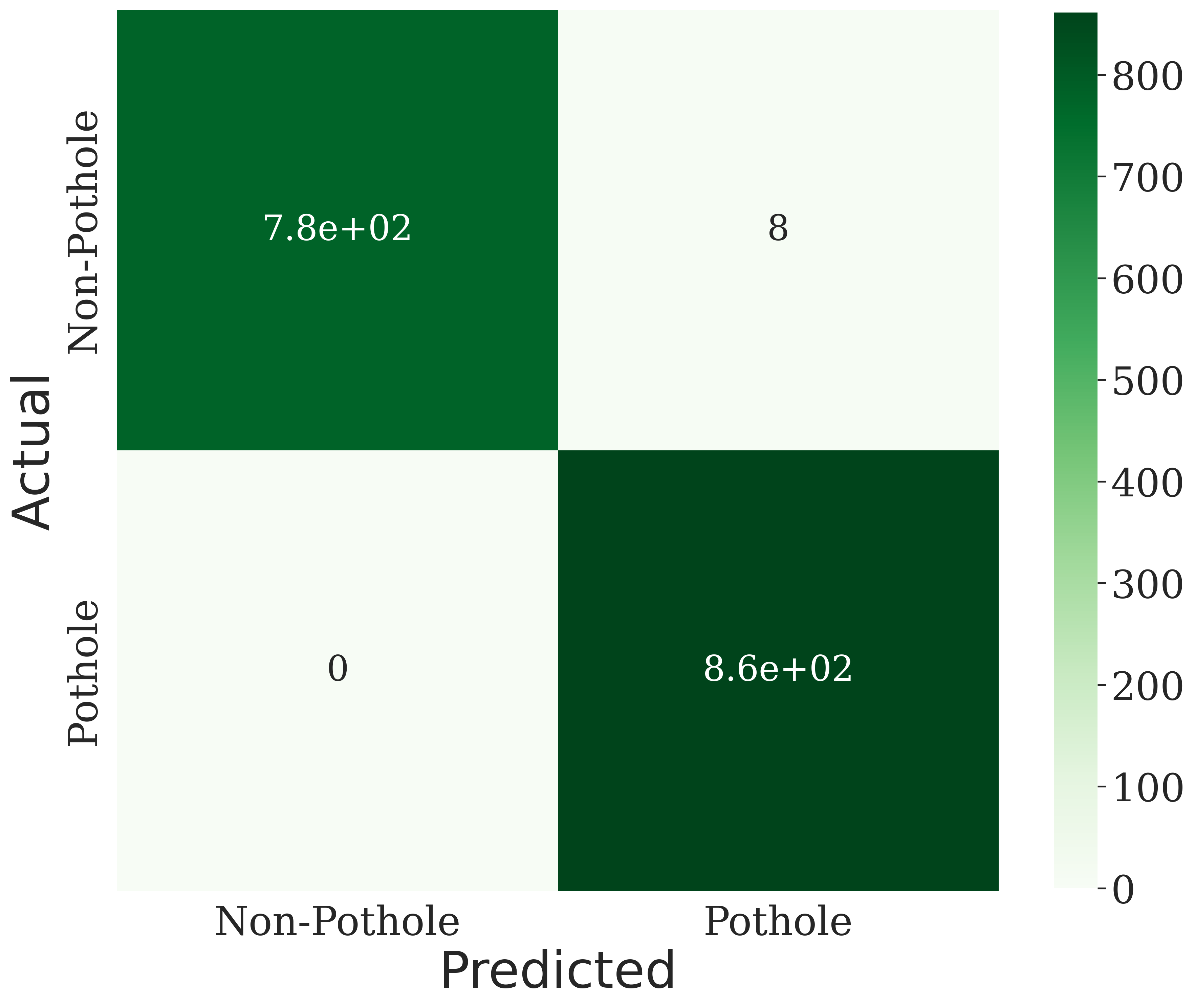}
        \caption{CCT (Augmented Dataset)}
        \label{fig:image2}
    \end{subfigure}
    \hspace{9mm} 
    \begin{subfigure}[b]{0.28\textwidth}
        \centering
        \includegraphics[width=\textwidth]{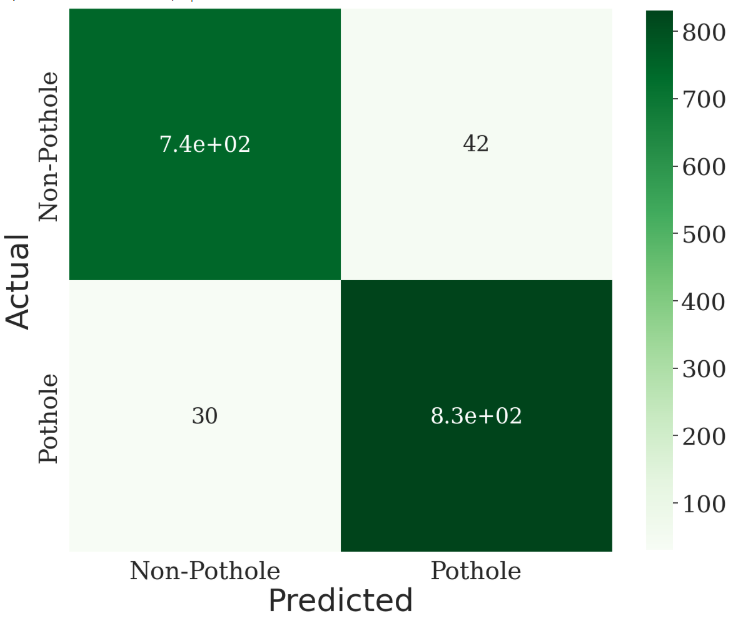}
        \caption{Swim Transformer (Augmented Dataset)}
        \label{fig:image2}
    \end{subfigure}
    \vspace{5mm}
    \begin{subfigure}[b]{0.28\textwidth}
        \centering
        \includegraphics[width=\textwidth]{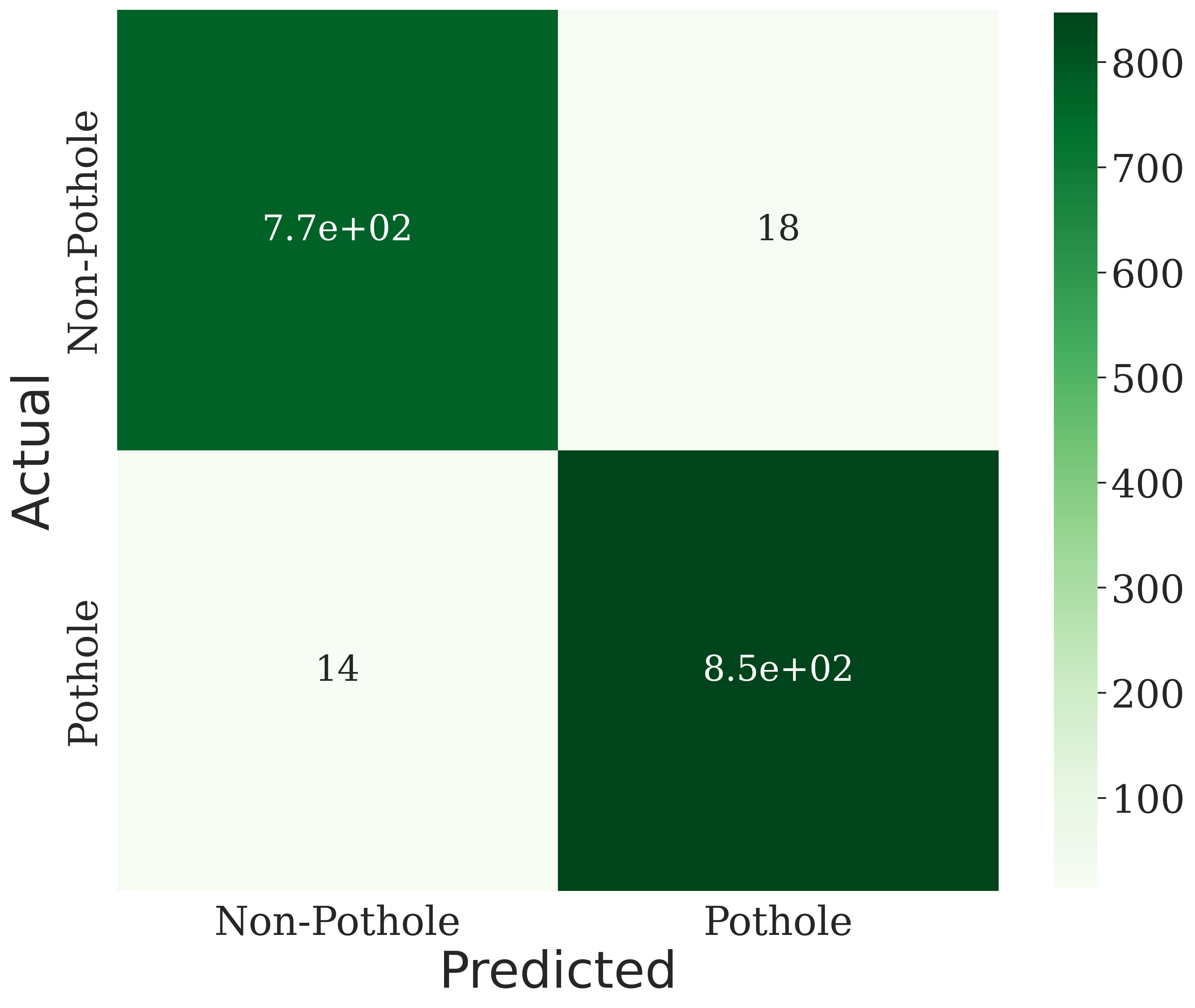}
        \caption{CNN (Augmented Dataset)}
        \label{fig:image2}
    \end{subfigure}
    \hspace{9mm} 
    \begin{subfigure}[b]{0.286\textwidth}
        \centering
        \includegraphics[width=\textwidth]{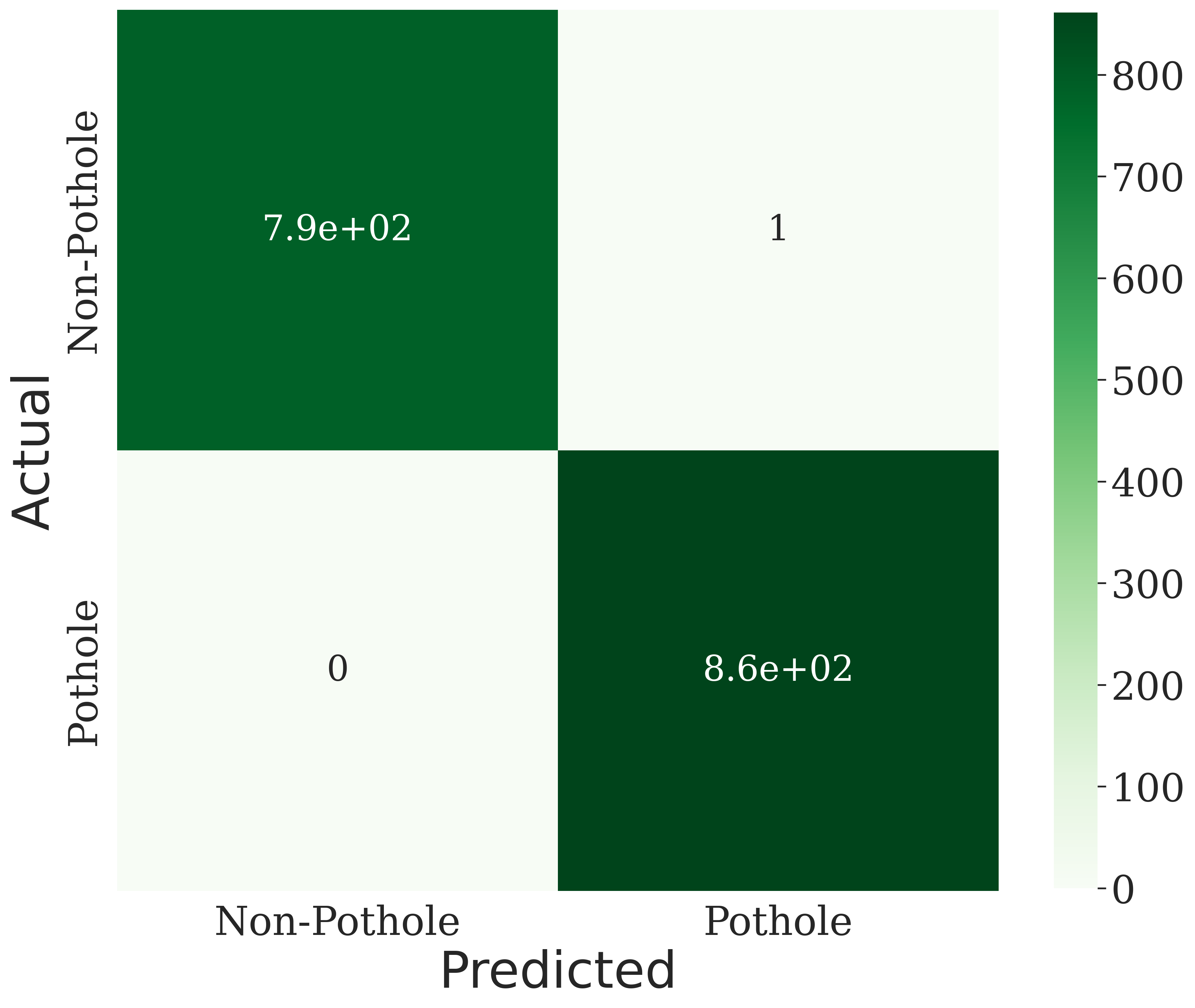}
        \caption{Convmixer (Augmented Dataset)}
        \label{fig:image2}
    \end{subfigure}
    \hspace{9mm} 
    \begin{subfigure}[b]{0.283\textwidth}
        \centering
        \includegraphics[width=\textwidth]{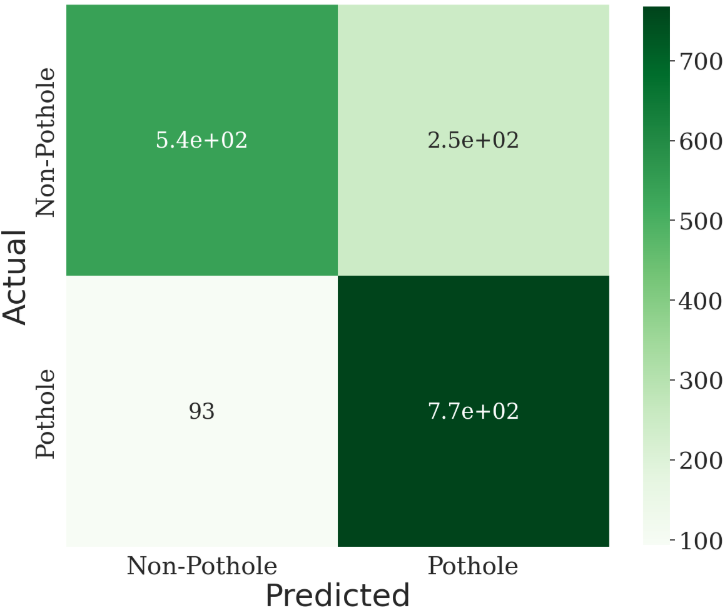}
        \caption{INN (Augmented Dataset)}
        \label{fig:image2}
    \end{subfigure}
    \vspace{1mm}

    \caption{Confusion Matrix (Classification Models)}
    \label{fig:sidebyside5}
\end{figure}

\begin{figure}[h!]
    \centering
    \begin{subfigure}[b]{0.22\textwidth}
        \centering
        \includegraphics[width=\textwidth]{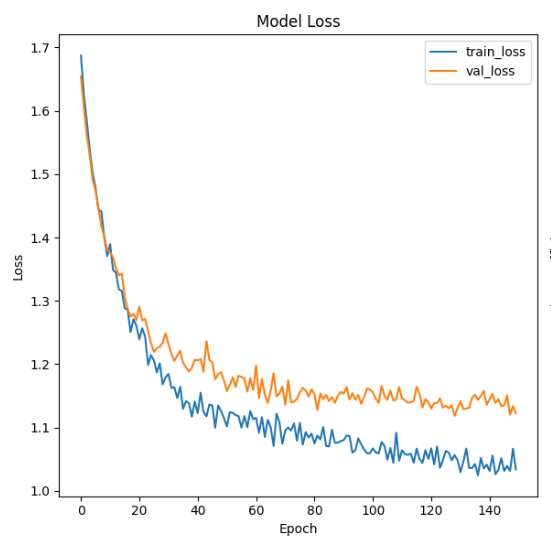}
        \caption{ U-Net (Original \newline Dataset)}
        \label{fig:u-net-original}
    \end{subfigure}
    \hspace{3mm}
    \begin{subfigure}[b]{0.22\textwidth}
        \centering
        \includegraphics[width=\textwidth]{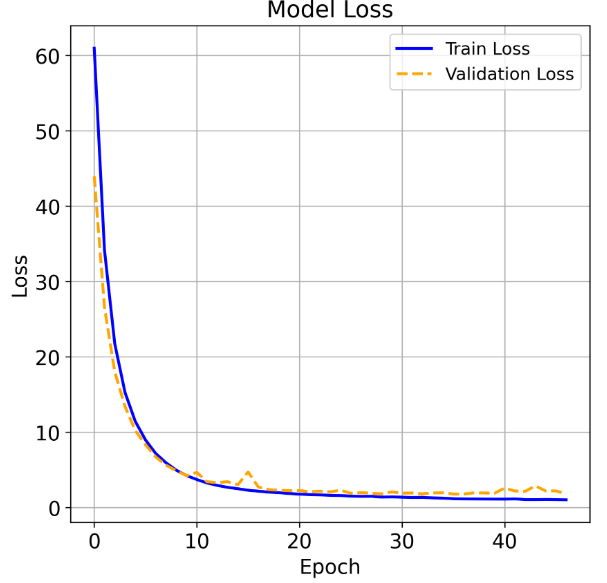}
        \caption{ResU-Net (Original Dataset)}
        \label{fig:u-net-augmented}
    \end{subfigure}
        \hspace{3mm}
    \begin{subfigure}[b]{0.22\textwidth}
        \centering
        \includegraphics[width=\textwidth]{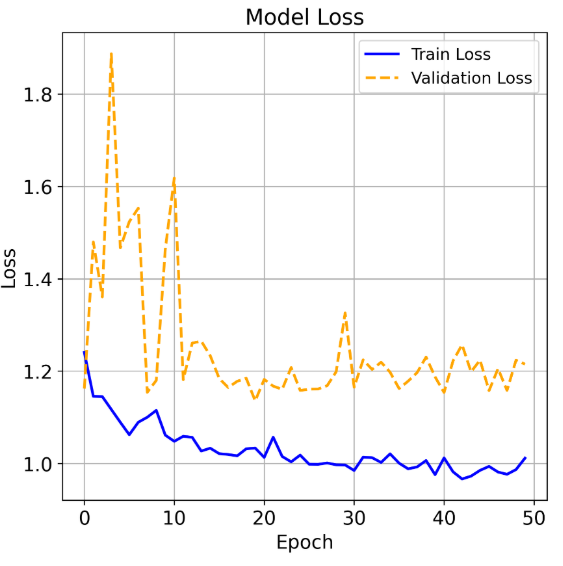}
        \caption{U-Net++ (Original Dataset)}
        \label{fig:u-net-augmented}
    \end{subfigure}
        \hspace{3mm}
    \begin{subfigure}[b]{0.22\textwidth}
        \centering
        \includegraphics[width=\textwidth]{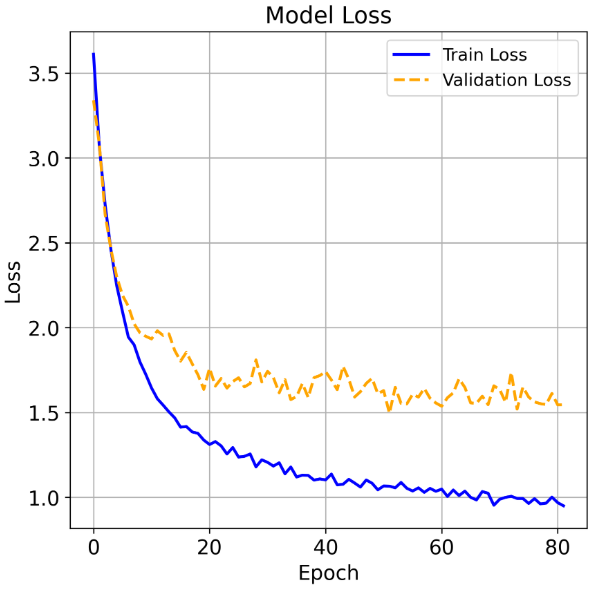}
        \caption{Attention U-Net (Original Dataset))}
        \label{fig:u-net-augmented}
    \end{subfigure}

    \vspace{1mm} 

    \begin{subfigure}[b]{0.22\textwidth}
        \centering
        \includegraphics[width=\textwidth]{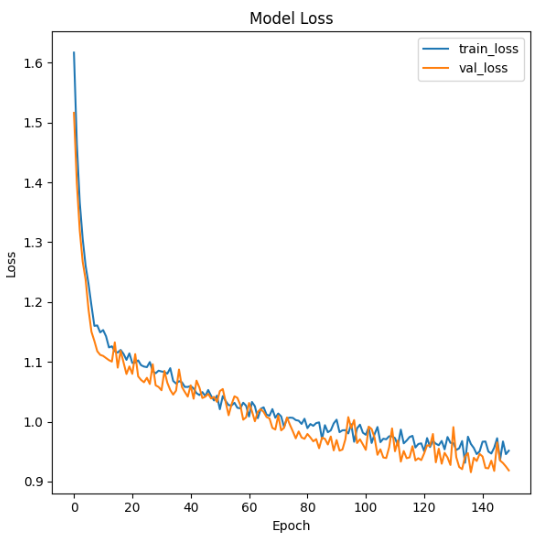}
        \caption{U-Net (Augmented Dataset)}
        \label{fig:resunet-original}
    \end{subfigure}
    \hspace{3mm}
    \begin{subfigure}[b]{0.22\textwidth}
        \centering
        \includegraphics[width=\textwidth]{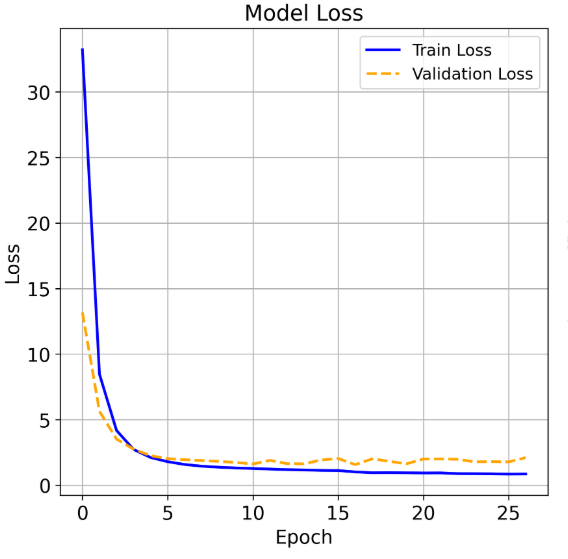}
        \caption{ResU-Net(Augmented Dataset)}
        \label{fig:resunet-augmented}
    \end{subfigure}
    \hspace{3mm}
    \begin{subfigure}[b]{0.22\textwidth}
        \centering
        \includegraphics[width=\textwidth]{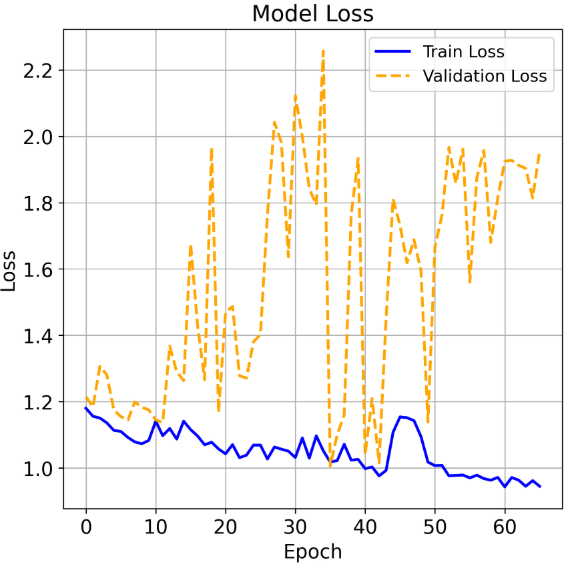}
        \caption{U-Net++ (Augmented Dataset)}
        \label{fig:unetpp-augmented}
    \end{subfigure}
    \hspace{3mm}
    \begin{subfigure}[b]{0.22\textwidth}
        \centering
        \includegraphics[width=\textwidth]{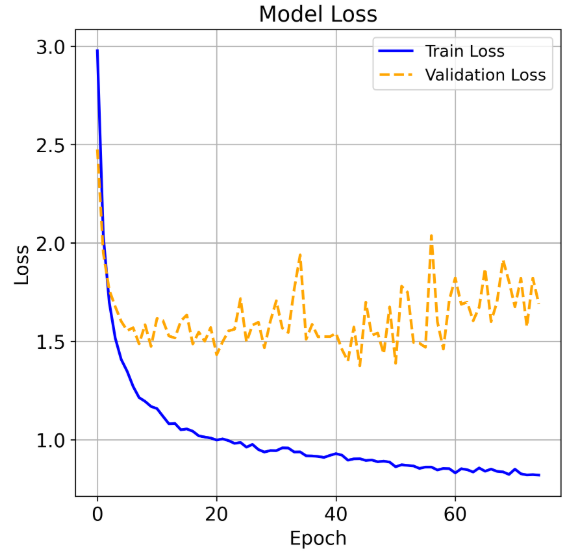}
        \caption{Attention U-Net (Augmented Dataset)}
        \label{fig:attention-augmented}
    \end{subfigure}
    \vspace{1mm}

    \caption{Loss Curves (Segmentaion Models)}
    \label{fig:segmentation-comparison6}
\end{figure}

VGG16, DenseNet201, Xception, CCT, and Convmixer exhibited accuracy and F1 scores over 99\% in the augmented dataset for classification, as illustrated in Table \ref{tab:example4}. Both Table \ref{tab:example4} and Table \ref{tab:example5} show that VGG16 outperformed all other models in classification for both datasets, followed by DenseNet201 and Xception. These three were pre-trained heavyweight models and maintained accuracy and F1 scores above 98\% on both datasets. ResNet50 posed to be the weak link among the pre-trained models, performing comparatively lower in the raw dataset and being the second worst in the augmented dataset. Also, ResNet50 was the only model in the list to have reduced performance upon augmentation. Among lightweight models, CCT was the overall best-performing model, having the highest accuracy and F1 score for the raw dataset, and the second highest accuracy and F1 score for the augmented dataset (among the lightweight models). Convmixer was the highest-improving model upon augmentation, performing the best among lightweight models in the augmented dataset while performing the worst for the raw dataset. Both CCT and Convmixer equated the heavyweight models in terms of performance numbers for the augmented dataset. Swin Transformer and CNN were also stable models and performed quite good over both datasets. INN had been the worst-performing lightweight model for both the datasets and the only model that can be inferred to have performed underwhelming overall. On the whole, augmentation enhanced the performance of all the models apart from ResNet50. Augmentation also improved the loss curves and confusion matrix of all the models in our testing which are shown in Figure \ref{fig:sidebyside5} and \ref{fig:segmentation-comparison6}. Nearly all the models performed decently while running on our dataset, indicating an up-to-the-mark quality of the collected Bangladeshi dataset. In comparison to the performance analysis encoded in the existing literature as shown in Table \ref{tab:related_work}, it is apparent that our proposed dataset generally performs on par or better compared to existing industrial and custom datasets for classification.

\begin{table}[ht]
\centering
\begin{tabular}{|l|l|l|l|l|}
\hline
Model & Accuracy (Raw Dataset) & Precision & Recall & F1 Score \\
\hline
INN & 46.06 & 0.23 & 0.42 & 0.35 \\
\hline
Densenet201 & 98.78 & 0.99 & 0.99 & 0.99 \\
\hline
Xception & 98.78 & 0.99 & 0.99 & 0.99 \\
\hline
VGG16 & 99.39 & 0.99 & 0.99 & 0.99 \\
\hline
ResNet50 & 92.72 & 0.93 & 0.93 & 0.93\\
\hline
CNN & 88.48 & 0.89 & 0.88 & 0.88 \\
\hline
ConvMixer & 46.06 & 0.21 & 0.46 & 0.29 \\
\hline
CCT & 90.90 & 0.91 & 0.91 & 0.91 \\
\hline
Swin Transformer & 89.09 & 0.90 & 0.89 & 0.89 \\
\hline
\end{tabular}
\caption{\label{tab:example4}Classification Results (Raw Dataset)}
\end{table}

\begin{table}[ht]
\centering
\begin{tabular}{|l|l|l|l|l|}
\hline
Model & Accuracy (Augmented Dataset) & Precision & Recall & F1 Score \\
\hline
INN & 79.429 & 0.80 & 0.79 & 0.79 \\
\hline
Densenet201 & 99.93 & 1 & 1 & 1 \\
\hline
Xception & 99.70 & 1 & 1 & 1 \\
\hline
VGG16 & 99.93 & 1 & 1 & 1 \\
\hline
ResNet50 & 89.68 & 0.94 & 0.94 & 0.93 \\
\hline
CNN & 98.05 & 0.98 & 0.98 & 0.98 \\
\hline
ConvMixer & 99.93 & 1 & 1 & 1 \\
\hline
CCT & 99.51 & 1 & 1 & 1 \\
\hline
Swin Transformer & 95.63 & 0.96 & 0.96 & 0.96 \\
\hline
\end{tabular}
\caption{\label{tab:example5}Classification Results (Augmented Dataset)}
\end{table}

\vspace{1.5mm}

For Segmentation, U-Net outperformed the other three models across both datasets, maintaining Dice coefficient and IoU scores of respectively 67.54\% and 59.39\% in the augmented dataset, as presented in Table \ref{tab:example7}, and 62.36\% and 51.43\% in the raw dataset illustrated in Table \ref{tab:example6}. U-Net’s dataset accuracy had also been the highest for both the datasets respectively 72.72\% for the augmented dataset and 62.23\% for the raw dataset. U-Net also maintained the best balance of precision and recall across both datasets, resulting in fewer false positives. Additionally, while ResU-Net had the highest Dice and IoU scores in the raw dataset, it had minimal improvements upon augmentation. Furthermore, Attention U-Net had the highest increment of Dice and IoU scores (10-15\%) upon augmentation. Augmentation overall increased the performance of the tested models and stabilized their loss curves, as shown in Figure \ref{fig:segmentation-comparison6}. On the whole, the augmentation results are acceptable for a real-world dataset, performing on par with the existing custom datasets tested across related literature, as referred in Table \ref{tab:related_work}. 

\vspace{3mm}

\begin{table}[ht]
\centering
\begin{tabular}{|l|l|l|l|l|l|}
\hline
Model & Accuracy (Raw Dataset) & Dice Coefficient & IoU & Precision & Recall \\
\hline
U-Net & 62.23 & 62.36 & 51.43 & 60.94 & 79.46 \\
\hline
ResU-Net & 54.74 & 64.92 & 53.32 & 54.02 & 99.79 \\
\hline
U-Net++ & 53.35 & 61.13 & 50.90 & 54.38 & 88.37 \\
\hline
Attention U-Net & 50.94 & 57.49 & 40.70 & 56.19 & 94.27 \\
\hline
\end{tabular}
\caption{\label{tab:example6}Segmentation Results (Raw Dataset)}
\end{table}

\begin{table}[ht]
\centering
\begin{tabular}{|l|l|l|l|l|l|}
\hline
Model & Accuracy (Augmented Dataset) & Dice Coefficient & IoU & Precision & Recall \\
\hline
U-Net & 72.72 & 67.54 & 59.39 & 77.48 & 71.72 \\
\hline
ResU-Net & 54.21 & 67.00 & 53.95 & 56.47 & 98.78 \\
\hline
U-Net++ & 53.79 & 64.78 & 53.81 & 59.33 & 78.51 \\
\hline
Attention U-Net & 53.63 & 67.49 & 49.28 & 63.30 & 85.85 \\
\hline
\end{tabular}
\caption{\label{tab:example7} Segmentation Results (Augmented Dataset)}
\end{table}


VGG16 excelled at classification due to its deep architecture consisting of 16 layers including 13 convolutional layers and 3 fully connected layers, allowing the model to learn more complex features from inputted pothole images\cite{Tammina2019}.  U-Net on the other hand excelled at segmentation despite being the simpler model among the 4 tested models. Models like U-Net++ and Attention U-Net generally excel on larger datasets and complex segmentation tasks. Pothole segmentation only involves two classes and does not fall under complex segmentation. It is evident from the research that, simpler models like U-Net are adequate for non-complex tasks like pothole segmentation, often outperforming the more complex models.


\vspace{3mm}

\section*{Discussion}

\subsubsection*{Insights:} It is evident that the proposed dataset performs competitively against the existing industrial and custom datasets utilized in the relevant literatures. The dataset excelled in classification with accuracy and f1 scores of over 99\% across five out of nine tested models. It also performed on par against the existing datasets in segmentation, reaching model dice scores up to 67.54\% and IoU scores up to 59.39\%. The careful data collection and pre-processing methodology associated with this dataset has contributed positively to its performance figures. Additionally, it was created complying with the ethical guidelines for data handling, supporting deletion or blurring of human faces and human-sensitive information, posing it as a credible public-use ready dataset. Along with that, the research indicates that simpler models like U-Net outperform the more complex models like U-Net++ and Attention U-Net in segmentation. The physical characteristics of potholes- shapes, textures, and edges seemingly align well with the assumptions of symmetry and locality encoded in U-Net\cite{Lin2017}. The pothole detection task does not benefit from the complex spatial relationships and hierarchical dependencies present in more complex models, as U-Net already captures the hierarchical structure inherent in the dataset, making additional complexity unnecessary. Complex models might be attempting to approximate features that are absent in the dataset, leading to inefficiency or overfitting. Our findings establish U-Net to be a more appropriate model for pothole segmentation.

\vspace{1mm}

\subsubsection*{Advantage of Augmentation:} This research additionally offers a more comprehensive, accurate, and a more mature showcase of the influence of data augmentation over deep learning models, tested across thirteen different architectures, combining both classification and segmentation results. Such large-scale research on the influence of augmentation has not been performed in any of the existing literature in our findings. As per our research, augmentation improved every performance metric of nearly every model in both testing methods. The performance metrics include- model accuracy, f1-scores, precision, recall, dice coefficient, and, IoU. Augmentation also stabilized the loss curves of all the models.  ResNet50 was the only model in our testing to possess slightly reduced performance figures after augmentation. Yet, the model’s loss curves were found to be stabilized in the augmented dataset.  

\vspace{1mm}

\subsubsection*{Efficiency of the Lightweight Models:} The research also provides an interesting perspective on the performance of lightweight deep learning models. By lightweight, we mean in terms of parameters. While having negligible parameters compared to the heavyweight models like ResNet50, VGG16, Xception and DenseNet201, the lightweight models like CCT and Convmixer performed nearly equal with surpassing accuracy scores over 99\% in the augmented dataset as presented in Table \ref{tab:example5} and Table \ref{tab:example8}. Lightweight models possess lower prediction times and require less computational resources, making them more appropriate for real-time detection tasks like pothole detection compared to heavyweight models. Utilization of lightweight models also entail compatibility with wide-ranging devices starting from low-power embedded devices to powerful data-center machines. This research overall presents a new outlook and consideration for the Lightweight models in computer vision tasks.


\vspace{3mm}
\section*{Limitations} 

The dataset samples were collected from two districts (Dhaka and Bogura) of Bangladesh among its sixty-four districts. Greater diversity of district data can reflect a more inclusive Bangladeshi scenario. Moreover, unpaved mud roads are not included in the dataset which is quite common in less developed parts of the country. In addition, the dataset samples were captured during winter for which there is absence of road condition during the rainy weather. To illustrate, potholes are commonly filled with water during the rainy season, changing their physical appearance. As such data is not present in the proposed dataset, so a model trained on this may not be able to accurately predict potholes filled with water. Similarly, the dataset does not involve road samples during snowfall due to the geographical limitation of Bangladesh. In addition, the dataset excludes night-time pictures due to being limited by the night-time capture quality of smartphone cameras. Consequently, the dataset is not suitable for training a model for pothole detection during nighttime. Lastly, we noticed that the models often detected shadows or areas of low brightness as potholes. We also noticed false positives for the sky and the surrounding environment.  


\begin{table}[ht]
\centering
\begin{tabular}{|l|l|l|}
\hline
\textbf{Classification or Segmentation} & \textbf{Model} & \textbf{Parameters} \\ 
\hline
\multirow{9}{*}{Classification} 
& INN & 1,573,967 \\ 
& CCT & 1,892,034 \\ 
& CNN & 2,938,051 \\ 
& Convmixer & 600,577 \\ 
& Swin Transformer & 627,905 \\ 
& VGG16 & 18,975,297 \\ 
& ResNet50 & 40,431,233 \\ 
& DenseNet201 & 23,542,593 \\ 
& Xception & 22,960,681 \\ 
\hline
\multirow{4}{*}{Segmentation} 
& U-Net & 105,473 \\ 
& U-Net++ & 9,042,177 \\ 
& ResU-Net & 32,462,849 \\ 
& Attention U-Net & 8,143,169 \\ 
\hline
\end{tabular}
\caption{\label{tab:example8} List of Parameters of Tested Models}
\end{table}
\vspace{1mm}
\section*{Conclusion and Future Work}

Deep-learning model architectures can efficiently predict potholes, preventing major damage. We have analyzed the performance of such 13 models for both classification and segmentation on our proposed dataset. A separate annotated dataset was created for segmentation. The raw datasets were then augmented and an overall comprehensive performance analysis was performed. We noticed that the models often falsely detected shadowy areas with low brightness as potholes. Furthermore, environmental elements such as trees and skies were often mislabeled as potholes. Thus, we observe via analysis that varying brightness values can have some degree of influence on predicting false positives. In addition, the lightweight models performed respectfully in comparison to the In future, we aim to diversify our dataset by adding other district roads of Bangladesh and samples containing rainy weather with water-filled potholes. We are also opting to test samples of unpaved roads as they are also quite common in the South-East Asian region. In addition, with proper funding to utilize professional cameras and private vehicles, we are willing to separately create and test a pothole dataset for a nighttime environment which we believe shall contribute significantly to this area of research. Along with that, we also aspire to work on a pothole dataset in foggy conditions.

\begin{table}[H] 
\centering

\begin{tabularx}{\textwidth}{|X|X|X|X|X|X|X|}
\hline
\textbf{Paper} & \textbf{Dataset} & \textbf{Best Model} & \textbf{Classification or Segmentation} & \textbf{Model Accuracy} & \textbf{Dice coefficient} & \textbf{IoU} \\ 
\hline
Ahmed et al., 2021 \cite{Ahmed2021} & Customized (MakeML, Roboflow \& some Self-collected) &  ResNet50 & Classification & 91.9\% & ... & ... \\ 
\hline
Arjapure et al., 2020 \cite{Arjapure2020} & Customized (Internet resources \& Self-collected) & DenseNet201 & Classification & 89.66\% & ... & ... \\ 
\hline
Ghosh et al.,2023 \cite{Susmita2023} & Self-Collected & Xception & Classification & 96.99\% &...&... \\
\hline
Parasnis et al., 2023 \cite{Guruprasad2023} & Collected from Kaggle & VGG16 & Classification & 95.5\% &...&...\\
\hline
Pramanik et al., 2021 \cite{Pramanik2021} & Self-collected & ResNet50  & Classification & 98.66\% &...&... \\ 
\hline
Zhang et al., 2023 \cite{Zhang2023} & Not mentioned & ResU-Net & Segmentation &...& 64.31\% & 73.24\% \\
\hline
Zhang et al., 2023 \cite{Zhang2023} & Not mentioned & U-Net &   Segmentation &...& 62.25\% & 72.13\% \\
\hline
Katsamenis et al., 2024 \cite{Katsamenis2024} & CrackMap & ResU-Net & Segmentation &&63.56\% & 46.58\% \\
\hline
Katsamenis et al., 2024 \cite{Katsamenis2024} & CrackMap & U-Net & Segmentation &...& 52.72\% & 35.79\% \\
\hline
Liu et al., 2023 \cite{Liu2023} & Customized (Self-collected \& Public datasets) & U-Net & Segmentation &...& 91.37\% & 84.79\% \\
\hline
Liu et al., 2023 \cite{Liu2023} & Customized (Self-collected \& Public datasets) & Attention U-Net & Segmentation &...&86.36\% & 76.7\% \\
\hline
Liu et al., 2023 \cite{Liu2023} & Customized (Self-collected \& Public datasets) & U-Net++ & Segmentation & ...& 92.78\% & 86.97\% \\
\hline
Our work & Self-collected  & VGG16 & Classification & 99.93\% & ... & ...\\
\hline
Our work & Self-collected  & U-Net & Segmentation & 72.72\% & 67.54 \% & 59.39\% \\
\hline
\end{tabularx}
\caption{Performance Across Related Literatures} 
\label{tab:related_work} 
\end{table}


\section*{Data Availability}
The current file encompasses all the data that were utilized, generated, or analyzed throughout the study.

\section*{Author Contributions}
A.F.P. and S.M.A. compiled and executed all experiments and coding for Classification. A.F.P. was responsible for coding and conducting experiments on Segmentation models. S.M.A. composed the manuscript and analyzed the results and findings. A.F.P., S.A.H., A.H.T., and M.A.S.K. collected the dataset. A.F.P., A.H.T, S.M.A., M.A.S.K, and S.A.H. annotated and augmented the dataset. S.M.A., A.H.T., M.A.S.K., and A.F.P. analyzed the existing literature. S.A.H., and M.A.S.K. produced all the figures utilized in the paper. A.F.P. and A.H.T. incorporated the manuscript on Latex. J.N. supervised the research. M.F.I. gave the research idea and co-supervised the research. J.N. and M.F.I. reviewed and corrected the manuscript.

\section*{Competing Interests}
The authors hereby declare that they have no financial, professional, or personal competing interests related to this research.

\end{document}